\pdfoutput=1

\documentclass[11pt]{article}

\usepackage[preprint]{acl}

\usepackage{times}
\usepackage{latexsym}

\usepackage[T1]{fontenc}

\usepackage[utf8]{inputenc}

\usepackage{microtype}

\usepackage{inconsolata}

\usepackage{times}
\usepackage{fancyhdr,graphicx,amsmath,amssymb}
\usepackage[ruled,vlined]{algorithm2e}

\usepackage{graphicx}
\usepackage{multirow}
\usepackage{xcolor}
\usepackage{comment}
\usepackage{ctable}

\title{Enhancing Data Privacy in Large Language Models through Private Association Editing}

\author{
\textbf{Davide Venditti$^{\textbf{*}1}$, 
Elena Sofia Ruzzetti$^{\textbf{*}1}$, 
Giancarlo A. Xompero$^{1,2}$}\\
\textbf{Cristina Giannone$^2$, Andrea Favalli$^2$, Raniero Romagnoli$^2$}\\
\textbf{Fabio Massimo Zanzotto$^1$} \\
$^1$University of Rome Tor Vergata, Italy \\
$^2$Almawave S.p.A., Via di Casal Boccone, 188-190 00137, Rome, IT
\\
\small{
\href{mailto:davide.venditti@uniroma2.it}{\color{black} \tt davide.venditti@uniroma2.it,}}
\small{
\href{mailto:elena.sofia.ruzzetti@uniroma2.it}{\color{black} \tt elena.sofia.ruzzetti@uniroma2.it}}
\\
\small{\href{mailto:fabio.massimo.zanzotto@uniroma2.it}{\color{black} \tt fabio.massimo.zanzotto@uniroma2.it}}
}

\begin{document}
\include{pythonlisting}
\maketitle

\def\thefootnote{*}\footnotetext{These authors contributed equally to this work}\def\thefootnote{\arabic{footnote}}

\begin{abstract}
Large language models (LLMs) require a significant redesign in solutions to preserve privacy in data-intensive applications due to their text-generation capabilities.
Indeed, LLMs tend to memorize and emit private information when maliciously prompted. 
In this paper, we introduce Private Association Editing (PAE) as a novel defense approach for private data leakage. PAE is designed to effectively remove Personally Identifiable Information (PII) without retraining the model.
Experimental results demonstrate the effectiveness of PAE with respect to alternative baseline methods. We believe PAE will serve as a critical tool in the ongoing effort to protect data privacy in LLMs, encouraging the development of safer models for real-world applications.
\end{abstract}

\section{Introduction}
Preserving privacy is the second major challenge for designers of data-intensive applications, and thus, it is a well-studied topic with well-established solutions. 
After meeting the applications' requirements, their focus immediately shifts to safeguarding users' privacy. Indeed,  
the association between Personally Identifiable Information (PII) and related data must be used following 
permissions granted by users and, possibly, enforced by law\footnote{EU with the General Data Protection Regulation (GDPR), the US with the Privacy Act, and China with the Personal Information Protection Law and Data Security Law}. 

Privacy is generally preserved 
by controlling access to the data repository and securing the communication channels, utilizing
privacy-by-design \citep{cavoukian2009privacy,schaar2010privacy,spiekermann2012challengesprivacy,cavoukian2012privacy} or 
cryptography \citep{ross2010visual,sun2011hcpp,barni2015privacy,abood2017investigation}.



\begin{figure}[h]
    \centering
    \includegraphics[width=7cm]{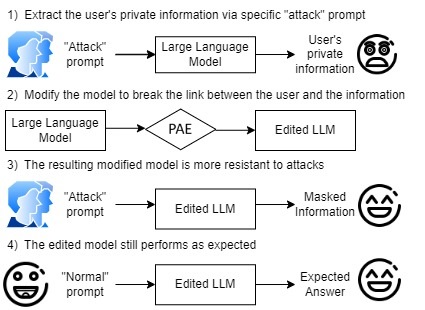}
    \caption{Preserving privacy for LLMs by using Private Association Editing}
    \label{fig:motivating_fancy_figure}
\end{figure}


Large language models (LLMs) require a significant shift in the solutions to preserve privacy 
due to their text-generation capabilities. Protecting training data and learned models in secure servers is not sufficient, as may be for machine learning classifiers. 
In fact, LLMs can unintentionally reveal associations between PII and sensitive information when prompted in specific ways \citep{carlini2019secret,carlini2023quantifying}. Their strength in abstraction and memorization \citep{Ozdayi2023,ranaldi-etal-2023-dark,ranaldi-etal-2023-precog}, paired with training on extensive web-scraped data as Common Crawl \citep{cc:Rana:2010:Common-Crawl-open-web-scale-crawl}, increases the likelihood of 
disclosing private information. Thus, designing privacy-preserving LLMs that can appropriately manage PII and sensitive data is a critical and compelling challenge that must be addressed \citep{brown2022does}.



In this paper, we propose a novel model to preserve privacy in LLMs: \textit{Private Association Editing} (PAE){, a ``one model, $k$ edits'' strategy} to remove memorized private information adjusting parameters of LLMs without re-training (see Fig.~\ref{fig:motivating_fancy_figure}). 
PAE is a model-editing privacy-preserving strategy based on the idea of \textit{breaking the association} between personal information and the identity of the person to whom it belongs by replacing the original information with masked -- but semantically equivalent -- information.
Inspired by recent model editing techniques \cite{meng2023locating,meng2023massediting}, PAE proposes two main innovations: the PAE cards and the PAE Regularization strategy. 
Experiments with GPT-J \cite{gpt-j} show that PAE outperforms alternative baseline methods in reducing privacy leaks without degrading the capabilities of LLMs to generate texts.   

The major contributions of the paper are:
\begin{itemize}
\vspace{-0.3cm}
  \item An innovative strategy to reduce privacy leak risks in LLMs: the PAE method that extends beyond factual editing approaches; 
\vspace{-0.3cm}
  \item Two important components of the PAE Method:  PAE Cards and PAE regularization;
\vspace{-0.3cm}
  \item The experimental analysis showing that PAE is an effective method to reduce privacy leaks and outperforms existing baseline methods.
\end{itemize}

\section{Background}

Large Language Models (LLMs) are prone to emit private information. Indeed, attacking LLMs to extract memorized private information is possible by using black-box access to language models.   
Training Data Extraction (TDE) is a technique to extract this private information \cite{carlini2021extracting}. It consists of querying the target model to force it to produce its own training data
and --as a result-- personal information inadvertently included in the training data like Twitter handles and email addresses \cite{carlini2021extracting}.
Those attacks are more effective if the private information to extract is preceded by the original training sequence in which it appeared \cite{carlini2023quantifying}. 
%
\citet{huang-etal-2022-large} demonstrated that conditioning a model with a prompt that is part of the training data can result in the leakage of personally identifiable information (PII), such as email addresses.
\citet{nasr2023scalable} revealed that \citet{carlini2021extracting} method is even more effective than previously expected:
by querying open-source models like GPT-Neo \cite{black2022gptneox20b} and Pythia \cite{biderman2023pythia}, they confirmed the success of the attack procedure using the training data solely for verification purposes.
%
Since these attacks require only black-box access to the model, closed models like GPT-3.5 and GPT-4 can be successfully attacked \cite{wang2024decodingtrust}. 



As personal information leakage from LLMs is a concrete possibility, {different strategies have been explored to avoid a model generating potentially harmful content. \citet{yao2024largelanguagemodelunlearning} propose an unlearning mechanism that requires only negative samples -- i.e., examples in which the model generates harmful content -- to stop the generation of undesirable outputs.
However, as the majority of machine unlearning approaches \cite{liu2024rethinkingmachineunlearninglarge}, it requires the definition of a retain set that contains samples used to preserve the utility of the model. 
Our aim is to modify only a batch of information without further training or additional data.
}

Model editing is a possible solution as opposed to an expensive remove-and-retrain strategy or {unlearning strategies}. 
Most of the previous work on model editing focuses on updating factual information.
\citet{mitchell2022memorybased} introduced a semi-parametric editing methodology, employing a retrieval-augmented counterfactual model. 
\citet{decao2021editing} 
edit factual knowledge within language models ensuring consistency across various formulations of facts. 
\citet{yao2023editing} introduced MEND on various datasets, demonstrating its ability to rapidly and effectively edit large-scale models’ behaviors without extensive retraining. 
Building on the idea that the linear layers in the Transformer architecture can be interpreted as key-value memories that store information \cite{geva-etal-2021-transformer},
ROME \cite{meng2023locating} and MEMIT \citep{meng2023massediting} demonstrate the ability to edit factual knowledge.
Since these methods can modify factual information memorized in LLMs, our goal is to exploit them to erase private information inadvertently ingested during training.

Similarly to the method defined in our paper, \citet{patil2023sensitive}
investigated model editing techniques to modify the information memorized in LLMs, concluding that 
information cannot be erased. 
In particular, they applied TDE attacks against the GPT-J \cite{gpt-j} model and demonstrated that in black-box access--performing attacks that also include paraphrases of the original prompt-- model editing cannot erase factual information memorized in GPT-J. 
Our setting is different: in fact, \citet{patil2023sensitive}
investigated the effectiveness of model editing only on factual information from sentences derived from Wikipedia and not directly present in the training data -- the Pile \cite{gao2020pile}.
By definition, the model under attack does not \textit{verbatim} memorize information that is not in training data: since the examples used by \citet{patil2023sensitive} are derived from Wikipedia and not included in the Pile, while the factual information they contain is memorized, they cannot be verbatim memorized. In our experiments, we directly study the effectiveness of model editing in deleting private information that is verbatim memorized 
rather than factual information.


\section{Attacking and Defending LLMs from Private Data Leakage with Private Association Editing}

Large Language Models (LLMs) have a tendency to emit private information memorized from their training data when fed with malicious prompts. 
In Training Data Extraction (TDE) attacks, if a model is prompted with a prefix encountered during training, it often completes it with the rest of the training sequence by producing verbatim private information \cite{carlini2023quantifying, huang-etal-2022-large}.

In this scenario, we propose a model to remove memorized Personally Identifiable Information (PII) from LLMs and, thus, reduce possible privacy leaks. Our procedure is extremely more versatile than erase-and-retrain and can be used in small batches of modification of an LLM.  It consists of 
three steps (see Fig.~\ref{fig:motivating_fancy_figure}): 
\begin{itemize}
    \vspace{-0.3cm}
    \item detecting the presence of memorized PII in \textit{pre-edit} LLMs performing black box TDE attacks (Sec.~\ref{sec:attacks});
    \vspace{-0.3cm}
    \item \emph{Private Association Editing} (PAE) to remove PII by editing parameters of LLMs obtaining \textit{post-edit} LLMs ~(Sec.~\ref{sec:defense}) 
    \vspace{-0.3cm}
    \item a final consistency check of \textit{post-edit} LLMs to assess that LLMs are not corrupted after PAE and behave similarly to \textit{pre-edit} LLMs~(Sec.~\ref{sec:consistency})
\end{itemize}



\subsection{Training Data Extraction Attacks to recover Sensitive Information}
\label{sec:attacks}

To detect the presence of memorized Personally Identifiable Information (PII) in LLMs, we follow the attack pipeline and attack prompts defined by \citet{huang-etal-2022-large}.
They defined two kinds of attacks depending on how information is stored and retrieved: 
(1) a model \textit{memorizes} personal information if there exists a prompt from the training data that leads the model to generate that information;
(2) in contrast, a model \textit{associates} an individual to its personal information if there exists a prompt not seen during training but containing a reference to an individual that leads to the generation of PII.
\citet{huang-etal-2022-large} already demonstrated that memorization is more common in LLM than association, showing that a model from the GPT-Neo 
family \cite{gpt-neo} can predict emails more accurately when conditioned with prompts from the training data rather than with unseen prompts.

We then analyze two attacking schemes: the Memorization attacks and the Association attacks.
In a \textbf{Memorization attack}, a model is fed with a prompt extracted from its pretraining data.
This prompt is the \textit{context} that precedes the private PII in the training data (see Appendix~\ref{app:attacks} for examples of these attacks).
The \textit{context} prompt is characterized by a length defined in terms of tokens.
In this attack, following \citet{huang-etal-2022-large}, we simulate that the attacker has 
more or less knowledge about the training data by conditioning the generation of the model to \textit{context} prompts of different lengths in terms of tokens. 
In the \textbf{Association Attack}, the model is instead fed with a prompt that contains an identifier of the person whose information is to be extracted but that does not exactly match the training data.
In particular, \citet{huang-etal-2022-large} defined four \textit{zero-shot} attack prompts, identified by letters from $a$ to $d$, all containing a reference to the identity -- the name -- of the PII owner.
Some examples of these attacks are reported in Appendix~\ref{app:attacks}.
The attack succeeds if, during the generation of the subsequent 
tokens, the model generates the target's private information, that is, the correct email address.
%
In both Memorization and Association attacks, the adversary in black-box access 
force the model to generate some PII regarding a person.

\subsection{Private Association Editing as Efficient Defense against Privacy Attacks}
\label{sec:defense}

{In order to protect people from privacy attacks performed on LLMs, we propose \textit{Private Association Editing} (PAE), an editing technique aiming at disrupting \textit{private associations}, i.e., associations between and individual and a PII 
included in the dataset used to train the LLM.} 
The technique proposed here is efficient since it allows the anonymization of private information directly into the model parameters, without 
retraining.

In this work, we define a private association as an association between the name of an individual and a PII that should not be revealed when interrogating the LLM.
The PAE cards -- for example \textit{The email address of \texttt{John Smith} is \texttt{john.smith@company.com}} -- describe this association between a person's name and its PII.
The PAE cards are the first component of our defense strategy.

Then, we propose the PAE Update Strategy on Model's Weights to mask the private information of individuals that has been inadvertently inserted into the training data.
PAE Update Strategy allows to substitute the PII with a semantically equivalent but anonymous value the target PII.

\paragraph{PAE Cards} 
The procedure to edit a private association uses PAE cards based on a modification card. The basic structure of a modification card is composed of a \texttt{prompt}, a \texttt{PII}, a \texttt{target}, and a \texttt{subject}. 
Our PAE cards specialize the modification card on a particular PII. 
We have defined two main types of PAE cards to mask the private information of users. The first type is called \textbf{explicit} because it frames the generation of the PII given a name of an individual as a task in zero-shot.
An explicit prompt in \textit{"Generate the email of \{name\}:"}
The second type is \textbf{implicit}, 
which frame the edit as a text generation task, given the person's name.
An example of an implicit prompt is \textit{"The mail address of \{name\} is"}.

\begin{table}[]
    \centering
    \resizebox{1.00\linewidth}{!}{
    \includegraphics{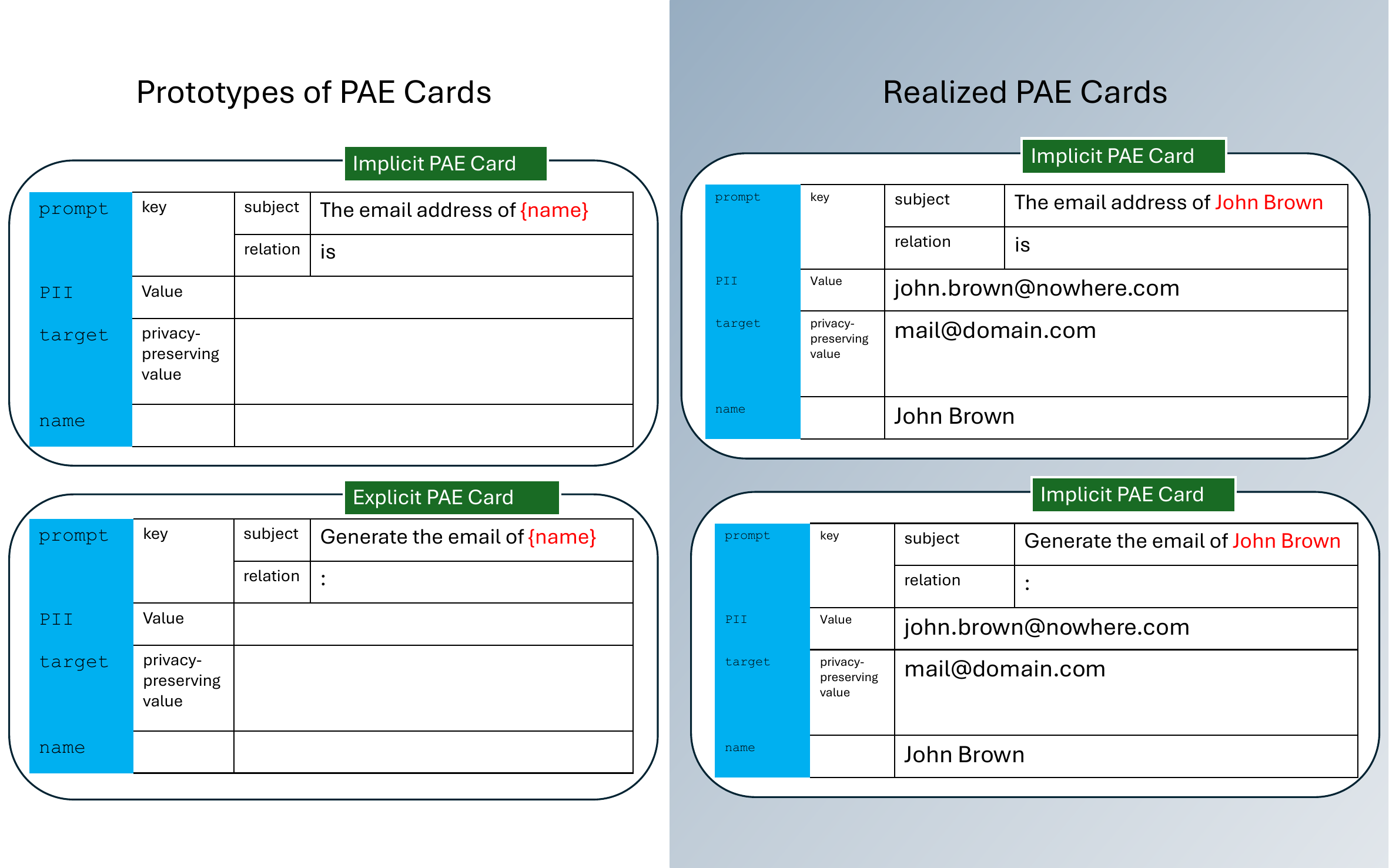}
    }
    \caption{Private Association Editing cards with two prototypes (Implicit and Explicit versions on email addresses) and a sample realization}
    \label{tab:PAE_for_email}
    \vspace{-0.3cm}
\end{table}

\paragraph{PAE Update Strategy on Model's Weights} 
PAE updates the Feed-Forward (FF) modules of target LLMs given the PAE Cards summarized in Table~\ref{tab:PAE_for_email}. In fact, studies suggest that the FF modules 
store information in the form of \textit{key-value} memories \cite{geva-etal-2021-transformer, meng2023locating, meng2023massediting}.
Thus, PAE edits matrices $W_0^l$, the last projection matrices in the FF module, to change memorized information: $\widehat{W_0^l} = W_0^l + \Delta$, where $l$ is the index layer, omitted when not necessary.

The update matrix $\Delta$ 
should break the association between
a \textit{key} encoding a \texttt{prompt} in Table~\ref{tab:PAE_for_email} 
and its corresponding \textit{value} encoding the \texttt{PII}.
To do so, PAE aims to substitute the current \textit{value} with a new, \textit{privacy-preserving value} anonymous \texttt{target}, that is semantically equivalent to the \texttt{PII} but does not violate any user privacy.

To determine $\Delta$, the target matrix $W_0$ should be written as the mapping between a set of keys $K_0$ and values $V_0$ learned during the pretraining phase $W_0 K_0 = V_0$ \cite{meng2023locating, meng2023massediting}. Hence, $\Delta$ matrix for PAE is defined as a function of the \textit{keys} $K \subset K_0$, the private \textit{values} $V \subset V_0$, and the new \textit{privacy-preserving values} $V^*$.

For PAE update strategy, we can frame the problem of finding the optimal update $\Delta$ to encode the privacy-preserving values $V^*$ imposing that the optimal post-update matrix $\widehat{W}^*$ -- defined as $\widehat{W}^* = W_0 + \Delta$ -- should be minimizing the following equation \cite{meng2023massediting}:
\begin{equation}
    \sum_{k\in K_0, v\in V_0}||\widehat{W}k - v||^{2}+\sum_{{k\in K, v^*\in V^*}}||\widehat{W}{k}-v^*||^{2}
\label{eq:maxim}
\end{equation}
{and keep the values $V^*$ = $\widehat{W}^*K$ similar to $V = W_0K$}.
Our update matrix $\Delta$ is then computed as:
\begin{equation}
    \Delta = \Lambda \otimes (V^* - W_0K)K^T(C_0 + KK^T)^{-1}
    \label{eq:delta_pae}
\end{equation}
where $\Lambda$ is a diagonal matrix defined as a function of the norm of $V$ and $V^*$ and $\otimes$ is the Hadamard product.
This equation is obtained as follows. As discussed by \citet{meng2023massediting}, the solution of Equation~\ref{eq:maxim} can be written in a closed form as: 
\begin{equation}
    \Delta' = (V^* - W_0K)K^T(C_0 + KK^T)^{-1}
    \label{eq:delta_memit}
\end{equation}
Intuitively, if $\Delta'$ is scaled by a constant multiplier $\lambda$ the post-edit model will be less consistent with respect to the pre-edit one as $\lambda$ increases and more consistent when $\lambda$ is closer to 0 (see 
Appendix~\ref{app:normalization}).
Then, in PAE, we introduce a mechanism for adjusting $\Delta'$ in function
of the relative weight of the new values $V^*$ and the old values {$V$}, {thus obtaining $\Delta=\Lambda \otimes \Delta'$}.
To design $\Lambda$, we analyze the $V^* - W_0K$ term of Equation~\ref{eq:delta_pae}: since by definition {$W_0K = V$}, 
the term can be rewritten as {$V^* - V$}.
Hence, the $i$-th row of the matrix {$V^* - V$} quantifies how different the \textit{privacy-preserving} value $v^*_i$ is from the corresponding $v_i$.
{We argue that the direction of this difference is important but the norm 
of the residuals should be comparable to the norm of the values before the update.
Then, after the update, the new values should be encoded in a similar way to what was done before the update.} 
Hence, we define the diagonal entry of $\Lambda$ term as:
\begin{equation}
\Lambda_{i,i} = \frac{||v_i||}{||v^*_i - v_i||+||v_i||}  
\label{lambda-regularization-update}
\end{equation}
to perform both the normalization and scale proportionally to {$||V||$}. 
The 
update rule in Equation~\ref{eq:delta_pae} allows to preserve users' privacy 
while maintaining the 
LLM utility (algorithm in Appendix \ref{app:algo}).



When using PAE, we adopt a strategy that we call ``one model, $k$ edits'': we are interested in subjecting the model to $k$ modifications at the time to comply to the real word scenario in which -- instead of performing single edits separately and recreating the model based on the post-edit weights obtained from the last edit every time -- $k$ different requests are addressed against a single model. As described in Section~\ref{sec:exp_setup}, the $k$ in PAE is not predetermined. 

By masking and anonymizing the email address, we make it more challenging for attackers to elicit specific private data from the model in response to particular prompts. This methodology effectively reduces the risk of sensitive information being inadvertently disclosed by the model.

\begin{table*}[]
\resizebox{\linewidth}{!}{
\begin{tabular}{|l|c|cc|cc|cc|}
\hline
\textbf{Editing} & \textbf{\small{LAMBADA}} & \multicolumn{2}{c|}{\textbf{Books3}}      & \multicolumn{2}{c|}{\textbf{Wikipedia}}   & \multicolumn{2}{c|}{\textbf{Pile-CC}}     \\
\textbf{Method}        & \textbf{Accuracy}         & \textbf{BLEU}                & \textbf{METEOR}              & \textbf{BLEU}                & \textbf{METEOR}             & \textbf{BLEU}                & \textbf{METEOR}              \\ \hline
FT                & 0.0 (-60\%)       & $63.4 (\pm 4.9) $  & $67.1 (\pm 4.8)$    & $63.0 (\pm 13.4)$   & $66.7 (\pm 10.9)$ & $60.9 (\pm 10.3)$ & $65.3 (\pm 7.7)$ \\ \hline
R-ROME            & 0.0 (-60\%)       & $63.3 (\pm 4.9) $  & $67.0 (\pm 4.8)$    & $63.0 (\pm 13.3)$   & $66.6 (\pm 10.9)$ & $60.9 (\pm 10.3)$ & $65.3 (\pm 7.7)$ \\ \hline
MEND              & 59.83 (-0.17\%)   & $91.6 (\pm 10.5) $ & $91.6 (\pm 10.6)$    & $89.3 (\pm 14.2)$   & $90.8 (\pm 12.5)$ & $91.5 (\pm 11.5)$ & $91.9 (\pm 11.3)$ \\ \hline
MEMIT \small{Implicit}    & 60.50  (+0.50\%)  & $86.6 (\pm 11.9) $ & $87.1 (\pm 12.3)$    & $87.5 (\pm 13.7)$   & $88.9 (\pm 12.6)$ & $89.1 (\pm 11.5)$ & $89.9 (\pm 11.2)$ \\
MEMIT \small{Explicit}    & 60.16  (+0.16\%)  & $89.0 (\pm 11.2) $ & $89.9 (\pm 10.6)$    & $88.3 (\pm 13.4)$   & $89.6 (\pm 12.1)$ & $88.8 (\pm 12.1)$ & $89.3 (\pm 11.8)$ \\ \hline 
PAE \small{Implicit}      & 60.50  (+0.50\%)  & $86.4 (\pm 11.2) $ & $87.3 (\pm 11.7)$    & $86.3 (\pm 13.0)$   & $87.6 (\pm 12.4)$ & $86.7 (\pm 13.) $ & $87.6 (\pm 12.4)$ \\
PAE \small{Explicit}      & 59.67  (-0.33\%)  & $86.2 (\pm 11.1) $ & $86.7 (\pm 11.5)$ 	  & $85.6 (\pm 13.3)$ 	& $87.5 (\pm 12.3)$ & $86.8 (\pm 11.6)$ & $87.5 (\pm 11.7)$ \\ \hline

\end{tabular}

}
\caption{Reliability of post-edited LLMs. In the first column, the LAMBADA accuracy score (for a comparison, the pre-edit accuracy score is $60\%$). To assess the similarity of the post-edit, we report BLEU and METEOR average scores on 300 examples drawn from 
Wikipedia, Books3, and Pile-CC Pile sub-datasets. The generations after PAE are similar to the ones from the pre-edit model. MEND maintains the higher similarity, but PAE is always comparable. Note that FT and R-ROME heavily reduce the model's capabilities (-60\% with respect to the baseline).} 
\label{tab:lm_eval}
\vspace{-0.3cm}
\end{table*}

\subsection{Evaluating Language Modeling Performance}
\label{sec:consistency}
The final step of the procedure 
is to investigate whether the LLM maintains its behavior in text generation. 
In fact, Model Editing techniques, in general, and PAE, in particular, may perturb the language model capabilities due to the intervention on the model parameters.
The LLM assessment procedure we describe in this Section aims to verify that the privacy-preserving language model is not a worse model than the original one. Since the models under investigation are foundational models, we focus on their language modeling capabilities rather than on an evaluation based on task performance.
If after the update the language model performs similarly to the pre-edit one, then also the performance on tasks will be similar.

We first introduce a metric for language model ability that can be used to assess the \textit{post-edit reliability}. 
The edit should cause no harm to the utility of the LM: to quantify this aspect, we adopt the LAMBADA \cite{paperno-etal-2016-lambada} benchmark.
LAMBADA measures the language modeling ability of a model calculating the accuracy the model has when asked to generate a missing target word from a passage. In the test split of the dataset, the missing word is always the last in the passage.
We will use 
the LAMBADA test set as the first indicator of the reliability of the edit.

However, we argue that the post-edit should not only demonstrate similar task performance but generate \textit{texts} as similar as possible to the pre-edit one: ideally,
we would like to have the post-update model indistinguishable from the pre-edit one.
The evaluation procedure we define is hence based on an automatic comparison between \textit{pre-edit} version LLM and a \textit{post-edit} version LLM.
The idea is to 
collect generations for a given set of prompts for \textit{pre-edit} LLM and \textit{post-edit} LLM. 
Then, these generations are compared with string-based similarity metrics, in particular BLEU and METEOR metrics. With these measures, we can automatically assess if \textit{pre-edit} LLM and \textit{post-edit} LLM behave similarly.
In Appendix~\ref{app:correlation_human}, we show that our method correlates with human judgements: for systems that achieve a high similarity in terms of BLEU or METEOR scores, annotators can only guess randomly whether the test examined is generated by the pre-edit or post-edit model.


\section{Experiments} 

\subsection{Experimental Setup}
\label{sec:exp_setup}
In this section, we discuss the parameters of our experiments: 
the analized LLM and related datasets, the application of PAE, and, finally, the set-up of the evaluation of the LLMs.
\paragraph{Analized LLM and related datasets}
In our experiments, we test the GPT-J model \cite{gpt-j}
that is designed to generate human-like text continuations from prompts: it is a large model, with 6 billion parameters, trained on the open dataset Pile \cite{gao2020pile}. 
The Pile is a 
large-scale text corpus that aggregates various sources, including books, articles, websites, and scientific papers.
The choice of the model and dataset is crucial since to effectively measure the performance of the attack, it is necessary to observe the training data \cite{carlini2021extracting, nasr2023scalable}.
However, this requirement is for evaluation purposes and does not limit the applicability of the PAE.
One of the constituent sub-datasets within The Pile is the Enron Emails \cite{klimt2004enron} corpus. This dataset contains text from approximately 150 users.
It includes a total of about 0.5 million email messages.
Its inclusion in the Pile mimics the inadvertent insertion into the training data of private information, in particular of PII-like email addresses:
the Enron dataset represents a natural starting point to test GPT-J memorization of PII.
We perform TDE discussed in Section~\ref{sec:attacks} to extract emails from the Enron corpus. We focus on greedy decoding since a preliminary study suggests no-difference between greedy and beam search attack accuracy (see Appendix~\ref{app:decoding_attacks}).

\paragraph{PAE and baselines application} 
PAE edits 
aim to cover the real-world scenario in which multiple privacy leakages are to be updated in a single edit, following a ``one model, $k$ edits'' philosophy. 
%
The are two distinct ways to apply model editing:
\textit{batch} editing that involves editing $k$ elements in an LLM simultaneously; \textit{sequential} editing focuses on editing $N$ elements within an LLM in a sequential way, 
with each edit on a subset of the $N$ elements.
A mixed approach that perform sequential edits of small batch sizes is closer to the real-world need to constantly update model parameters, with privacy leakages may be discovered over time.
{PAE can effectively preserve the privacy of users both with a small number of large batch edits and with a larger number of smaller batch edits in a sequential fashion.
We adopt a large batch size with $k=N$ as this is in principle the safest approach since the post-edit parameters are directly the pre-edit ones.
Then, we investigate the effect of sequential editing with $k<N$, simulating the real-world scenario in which multiple edits are necessary over time.
PAE is compared with a number of baselines:
we adopt a naive Fine-Tuning (FT) approach to instruct the model to generate the new \texttt{target} in place of the original \texttt{PII};
ROME \cite{meng2023locating} in its R-ROME implementation \cite{gupta2024rebuildingromeresolving} is also tested with fully sequential scheme;
MEND \cite{yao2023editing} requires a meta-training to define the update and in our experiments we adopt the same model as in the original paper, while we apply it on the PAE cards.
Finally, MEMIT \cite{meng2023massediting} is applied as baseline itself.
The PAE Implicit Card is applied as an edit prompt for all the baselines. MEMIT which is the most similar to PAE is also tested against PAE Explicit Cards.
We then show that PAE leads to a decrease in privacy leakage compared to baseline methods. 

\begin{table*}[]
\resizebox{\linewidth}{!}{

\begin{tabular}{|lr|ccc|cccc|}
\hline
\multicolumn{2}{|c|}{\multirow{2}{*}{\textbf{Editing Method}}} & \multicolumn{3}{c|}{\textbf{Memorization Attacks}}                & \multicolumn{4}{c|}{\textbf{Association Attacks}}                                         \\
\multicolumn{2}{|l|}{}                                         & \textit{context 50} & \textit{context 100} & \textit{context 200} & \textit{zero shot a} & \textit{zero shot b} & \textit{zero shot c} & \textit{zero shot d} \\ \hline
Pre-edit                 & accuracy                 & 12.49               & 16.23                & 18.2                 & 0.16                 & 0.06                 & 0.8                  & 2.1                  \\
\multicolumn{2}{|r|}{\textit{\# leak/ \#gen}}                  & \textit{353/2827}   & \textit{476/2932}    & \textit{537/2951}    & \textit{5/3130}      & \textit{2/3229}      & \textit{26/3234}     & \textit{68/3237}     \\ \hline
MEND                                & accuracy                 & 8.36                & 10.97                & 12.62                & 0                    & 0                    & 0.4                  & \textbf{1.05}        \\
\multicolumn{2}{|r|}{\textit{\# leak/ \#gen}}                  & \textit{233/2786}   & \textit{317/2891}    & \textit{367/2909}    & \textit{0/3128}      & \textit{0/3226}      & \textit{13/3222}     & \textit{34/3234}     \\ \hline
MEMIT \small{Implicit}                      & accuracy                 & 7.37                & 10.54                & 12.83                & 0.03                 & 0                    & 0.4                  & 1.48                 \\
\multicolumn{2}{|r|}{\textit{\# leak/ \#gen}}                  & \textit{203/2753}   & \textit{301/2856}    & \textit{368/2868}    & \textit{1/3109}      & \textit{0/3218}      & \textit{13/3222}     & \textit{48/3234}     \\
MEMIT \small{Explicit}                      & accuracy                 & 8.03                & 11.41                & 13.29                & 0.03                 & 0                    & 0.56                 & 1.21                 \\
\multicolumn{2}{|r|}{\textit{\# leak/ \#gen}}                  & \textit{221/2751}   & \textit{326/2858}    & \textit{383/2882}    & \textit{1/3102}      & \textit{0/3219}      & \textit{18/3229}     & \textit{39/3232}     \\ \hline
PAE \small{Implicit}                        & accuracy                 & \textbf{6.14}       & \textbf{8.85}        & \textbf{11.13}       & \textbf{0}           & \textbf{0}           & \textbf{0.37}        & 1.43                 \\
\multicolumn{2}{|r|}{\textit{\# leak/ \#gen}}                  & \textit{173/2816}   & \textit{256/2893}    & \textit{325/2921}    & \textit{0/3171}      & \textit{0/3230}      & \textit{12/3220}     & \textit{46/3227}     \\
PAE \small{Explicit}                        & accuracy                 & 7.03                & 9.69                 & 11.6                 & 0.03                 & 0                    & 0.5                  & \textbf{1.08}        \\
\multicolumn{2}{|r|}{\textit{\# leak/ \#gen}}                  & \textit{191/2718}   & \textit{276/2848}    & \textit{335/2887}    & \textit{1/3032}      & \textit{0/3215}      & \textit{16/3226}     & \textit{35/3230}     \\ \hline
\end{tabular}

}
\caption{Effectiveness of removing Private Personal Information from LLMs with different methods. Comparison of Training Data Extraction attacks accuracy with \textit{Pre-edit} model and after the editing via \textit{PAE} and baseline methods.}
\label{tab:results}
\vspace{-0.5cm}
\end{table*}
\vspace{-0.2cm}
\paragraph{Evaluation of post-edit LLM}
We will adopt LAMBADA as a first indicator of model editing technique being reliable. If a model editing technique can preserve accuracy on this task (as we discussed in Section~\ref{sec:consistency}) then we claim that the editing is reliable. We will report the result on 600 examples drawn from the LAMBADA test set.

Moreover, we introduce an additional set of experiments to ensure that \textit{texts} generated after the editing are close to the one generated by the pre-edit model.
We measure the difference in generations for the pre-trained GPT-J model and the post-edit version by generating a 50 token long paragraph starting from {a total of 300 examples from the Pile, obtained by extracting 100 examples from its Book3 \cite{rae2022scaling}, Wikipedia and Pile-CC sub-dataset.
We prompted the post-edit models and the pre-edit one with 100 tokens of the 300 randomly selected examples, and we evaluated how similar the generations are by measuring their overlap. The higher the similarity, the lower the influence of PAE on the model performance.}
Evaluation metrics are ROUGE and METEOR scores.


\subsection{Results and Discussion}
\label{sec:res}

\paragraph{LLMs leak Private Information}
Since LLMs tend to leak training data, we aim to quantify the amount of private information that can be retrieved from the pre-trained GPT-J. Unfortunately, GPT-J makes no exception to the trend noticed by \citet{huang-etal-2022-large} for the GPT-Neo models. In fact, this model also tends to generate PII.

In Table~\ref{tab:results}, it is possible to observe that Training Data Extraction Attacks that are based on Memorization are particularly effective against the GPT-J model: on average, the model tends to accurately predict the mail observed during training the $16\%$ of the times. 
It is worth noting the scale of the leakage: the model is originally prompted with 3238 examples.
On average, $455.3$ emails are correctly generated by those attacks: the privacy of a large number of people is threatened.

Moreover, as the attacker gets more information, the accuracy of the attacks gets higher: the accuracy of the attacks strongly depends on the length of the prompt. 
In fact, the lower accuracy 
that can be registered in Memorization Attacks is $12.49\%$: the model, in that case, is fed with a \textit{context} prompt that is $50$ tokens long.
However, when the \textit{context} prompt given to the model is composed of $200$ tokens, the accuracy of the attack peaks at $18.2\%$. 

The accuracy of the Association Attacks is much more modest. The results of those attacks against GPT-J model exhibit similar patterns to the one observed by \citet{huang-etal-2022-large} against the GPT-Neo.
The larger number of email addresses leaked by those kinds of attacks is $68$, a modest number compared to the accuracy obtained in the Memorization Attacks.
However, in an adversarial scenario, even low accuracy may cause harm: 
we will demonstrate the efficacy of PAE against both types of attacks.

\vspace{-0.5em}
\paragraph{PAE preserves the LM Capabilities}
Before evaluating the effect of the edits to preserve privacy, we focus on identifying which edit strategies maintain the model's utility and, hence, whether they are reliable or not.

The LAMBADA accuracy gives an intuition of which methods are more reliable: most of the methods achieve similar accuracy. In particular, the majority of the methods achieve performance close with respect to the pre-edit model, which achieves an accuracy of $60\%$. However, the FT and R-ROME methods both disrupt the LM ability of the model. These results already allow us to identify which methods cause a model collapse with the high number of edits and align with previous findings \cite{gupta2024modeleditingscaleleads}.

When evaluating the similarity of texts genenerated by different model editing techniques with respect to the pre-edit model
-- both according to the BLEU metric and to METEOR -- the systems generate 
very similar paragraphs when prompted with the same tokens.
While the update method that always has a closer similarity to the pre-edit model is MEND, PAE always has comparable similarity scores. In fact, the differences are not statistically significant, as the t-test fails with a p-value greater than $0.05$.
The same trend can be observed comparing PAE and the MEMIT update strategy.
It is also possible to notice the decreased utility after the FT and R-ROME: both the BLEU and METEOR scores suggest that the model has a completely different behavior.
%

This evaluation procedure can attest that PAE, MEND, and MEMIT are applicable because leave the capabilities of the language model intact. In the next paragraph we will discuss how PAE is more effective than the MEND and MEMIT baselines in in editing private information.

\vspace{-0.5em}
\paragraph{PAE in batch editing Preserves Privacy}
In Table~\ref{tab:results}, it is possible to observe the reduced effectiveness of Memorization and Association attacks after the GPT-J model has undergone an editing process 
with PAE, demonstrating its efficacy 
against all the attacks, also the most informed ones.
We argue that PAE edits are effective since they can reduce the leakage of private information, regardless of the nature of the attack.

PAE is an effective solution against Memorization Attacks. In particular, the accuracy of the attacks steadily decreases in each configuration. The average drop in attack accuracy after a PAE Implicit edit is {$6.93\%$}, and {$6.2\%$} after a {PAE Explicit edit}: this means that PAE {Implicit} is able to modify the model parameters so that, on average, the {$45.6\%$} of the previously predicted email addresses are no longer verbatim generated by the model using the implicit defense strategy, and {$41.8\%$ with the explicit one}.
Against attacks with \textit{context} prompt of \textit{200} tokens, PAE effectiveness peaks with $60.52\%$ of the email addresses leaked from the pre-edit model being anonymized.
While being always comparable, the PAE Implicit strategy is more effective than the PAE Explicit one, with the exception of the Association Attacks in \textit{zero shot d} configuration.

In absolute number of leakage, more informed \textit{context} prompts are still challenging; however, it must be observed that the accuracy of the most robust attack, with \textit{context} prompts of 200 tokens, 
after the application of PAE Implicit is similar to the accuracy of the less robust Memorization attack with a \textit{context} of \textit{50} tokens.
This analysis shows that PAE can help in protecting privacy.
In addition, PAE is more effective than the MEND and MEMIT baselines.
Those strategies make the model more robust against any attack configuration, but are less effective than PAE: they manage to achieve, respectively, a masking of {$36.9\%$} and {$33\%$} of previously leaked emails.
The FT and R-ROME baseline, as discussed in the previous paragraph, are not reliable since they disrupt the model utility (their effect on attacks is discussed in Appendix~\ref{app:ft_rome}).


Post-edit results show a significant reduction in the effectiveness also of Association attacks. This reduction is particularly notable in scenarios where the number of leaked emails drops close to zero.
Crucially, while not perfect, the PAE edits -- both Implicit and Explicit -- always cause an increase in privacy protection since they reduce the number of emails correctly leaked by Association Attacks.
However, it is crucial to consider the originally leaked emails when interpreting post-edit results. While a reduction to near-zero leakage is impressive, the impact is more pronounced when starting from a higher number of pre-edit leaks.

\vspace{-0.5em}
\paragraph{PAE ``one model, $k$ edits'' is flexible}
\begin{figure}[t]
    \centering
    \includegraphics[width=7cm]{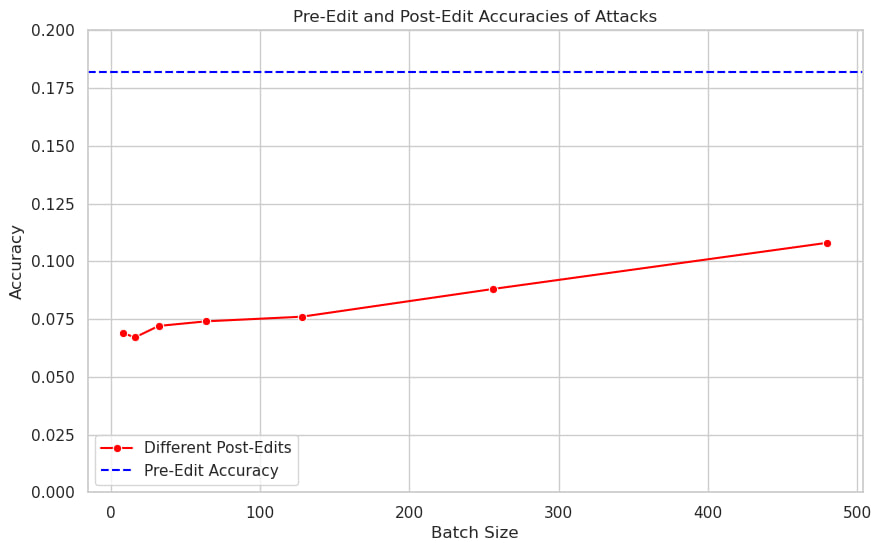}
    \caption{Memorization Attack against sequentially post-edit models. The smaller the batch size $k$, the larger the number of sequential updates necessary to edit all the private email addresses leaked by the original model.
    }
    \label{fig:seqedit}
    \vspace{-0.3cm}
\end{figure}
Finally, in Figure~\ref{fig:seqedit} it is possible to notice that 
{the ``one model $k$ edits approach'' is flexible and can be applied with different $k$, successfully mixing batch and sequential editing to preserve users' privacy.}
In these experiments, we perform sequential edits of the GPT-J model, varying the number of email addresses anonymized per edit, varying $k$ from $8$ to $480$. We indicate the number of anonymized emails per edit as batch size $k$: with $k \ll N$, we mimic the real-world scenario of updating a model each time a privacy leak is detected.
To understand whether sequential editing has a negative impact on the effectiveness of the edit, we evaluate the effectiveness of PAE for each of the batch sizes in the Memorization Attack with the more effective of the prompts ($|p_M|=200$). The results refer to a model post-edit with "implicit" PAE.
The accuracy of the edit is rather stable and similar to the results obtained in the batch editing scenario.
{Also, the underlying language model is not negatively affected by the different $k$, as reported in Appendix~\ref{app:sequential_evaluation}.
Those results also confirm the applicability of PAE in sequential editing, demonstrating the validity of the ``one model, $k$ edits'' approach.
}

\vspace{-0.2cm}
\section{Conclusion}
\vspace{-0.2cm}
In this paper, we address the critical issue of private data leakage in Large Language Models (LLMs) due to their tendency to memorize training data. We propose Private Association Editing (PAE), a novel defense mechanism that effectively removes Personally Identifiable Information (PII) from LLMs without requiring retraining.
Our methodology involves a three-step procedure: 
detecting memorized PII via Training Data Extraction (TDE) attacks,
applying PAE to preserve users' privacy,
and ensuring consistency in the post-edit LLMs. 
The PAE method is more effective than a number of baselines in preserving users' privacy.

Our experiments demonstrate that the PAE approach is both effective and efficient in mitigating the risk of private data leakage. 
We believe PAE will be a valuable tool in the ongoing effort to protect data privacy in LLMs and encourage its adoption to prevent potential privacy violations.

\section*{Limitations}
We outline some limitations and possible directions for future research in enhancing data privacy in Large Language Models (LLMs).

As the landscape of LLMs evolves, it may be useful to extend the Private Association Editing (PAE) mechanism to accommodate new types of models and data.
Currently, we apply our proposed PAE method on a limited set of LLMs. A possible extension could involve testing and refining PAE across a broader spectrum of LLM architectures and training datasets. However, research in this direction requires open model and open training data.
Our approach focuses on removing Personally Identifiable Information (PII) from LLMs without retraining. However, this method might not address all types of sensitive data. Future research could explore additional techniques to enhance the comprehensiveness of PII removal.
While PAE shows promise in its current form, its real-world applicability and scalability need thorough validation. 
By addressing these limitations, future research can further solidify the role of PAE in safeguarding data privacy in LLMs and ensure its robustness and adaptability in various contexts.

\bibliography{anthology,custom}

\begin{thebibliography}{40}
\providecommand{\natexlab}[1]{#1}

\bibitem[{Abood et~al.(2017)Abood, Elsadd, and Guirguis}]{abood2017investigation}
Omar~G Abood, Mahmoud~A Elsadd, and Shawkat~K Guirguis. 2017.
\newblock Investigation of cryptography algorithms used for security and privacy protection in smart grid.
\newblock In \emph{2017 Nineteenth International Middle East Power Systems Conference (MEPCON)}, pages 644--649. IEEE.

\bibitem[{Barni et~al.(2015)Barni, Droandi, and Lazzeretti}]{barni2015privacy}
Mauro Barni, Giulia Droandi, and Riccardo Lazzeretti. 2015.
\newblock Privacy protection in biometric-based recognition systems: A marriage between cryptography and signal processing.
\newblock \emph{IEEE Signal Processing Magazine}, 32(5):66--76.

\bibitem[{Biderman et~al.(2023)Biderman, Schoelkopf, Anthony, Bradley, O'Brien, Hallahan, Khan, Purohit, Prashanth, Raff, Skowron, Sutawika, and van~der Wal}]{biderman2023pythia}
Stella Biderman, Hailey Schoelkopf, Quentin Anthony, Herbie Bradley, Kyle O'Brien, Eric Hallahan, Mohammad~Aflah Khan, Shivanshu Purohit, USVSN~Sai Prashanth, Edward Raff, Aviya Skowron, Lintang Sutawika, and Oskar van~der Wal. 2023.
\newblock \href {https://arxiv.org/abs/2304.01373} {Pythia: A suite for analyzing large language models across training and scaling}.
\newblock \emph{Preprint}, arXiv:2304.01373.

\bibitem[{Black et~al.(2022)Black, Biderman, Hallahan, Anthony, Gao, Golding, He, Leahy, McDonell, Phang, Pieler, Prashanth, Purohit, Reynolds, Tow, Wang, and Weinbach}]{black2022gptneox20b}
Sid Black, Stella Biderman, Eric Hallahan, Quentin Anthony, Leo Gao, Laurence Golding, Horace He, Connor Leahy, Kyle McDonell, Jason Phang, Michael Pieler, USVSN~Sai Prashanth, Shivanshu Purohit, Laria Reynolds, Jonathan Tow, Ben Wang, and Samuel Weinbach. 2022.
\newblock \href {https://arxiv.org/abs/2204.06745} {Gpt-neox-20b: An open-source autoregressive language model}.
\newblock \emph{Preprint}, arXiv:2204.06745.

\bibitem[{Black et~al.(2021)Black, Gao, Wang, Leahy, and Biderman}]{gpt-neo}
Sid Black, Leo Gao, Phil Wang, Connor Leahy, and Stella Biderman. 2021.
\newblock \href {https://doi.org/10.5281/zenodo.5297715} {{GPT-Neo: Large Scale Autoregressive Language Modeling with Mesh-Tensorflow}}.
\newblock {If you use this software, please cite it using these metadata.}

\bibitem[{Brown et~al.(2022)Brown, Lee, Mireshghallah, Shokri, and Tram{\`e}r}]{brown2022does}
Hannah Brown, Katherine Lee, Fatemehsadat Mireshghallah, Reza Shokri, and Florian Tram{\`e}r. 2022.
\newblock What does it mean for a language model to preserve privacy?
\newblock In \emph{Proceedings of the 2022 ACM Conference on Fairness, Accountability, and Transparency}, pages 2280--2292.

\bibitem[{Cao et~al.(2021)Cao, Aziz, and Titov}]{decao2021editing}
Nicola~De Cao, Wilker Aziz, and Ivan Titov. 2021.
\newblock \href {https://arxiv.org/abs/2104.08164} {Editing factual knowledge in language models}.
\newblock \emph{Preprint}, arXiv:2104.08164.

\bibitem[{Carlini et~al.(2023)Carlini, Ippolito, Jagielski, Lee, Tramer, and Zhang}]{carlini2023quantifying}
Nicholas Carlini, Daphne Ippolito, Matthew Jagielski, Katherine Lee, Florian Tramer, and Chiyuan Zhang. 2023.
\newblock \href {https://arxiv.org/abs/2202.07646} {Quantifying memorization across neural language models}.
\newblock \emph{Preprint}, arXiv:2202.07646.

\bibitem[{Carlini et~al.(2019)Carlini, Liu, Úlfar Erlingsson, Kos, and Song}]{carlini2019secret}
Nicholas Carlini, Chang Liu, Úlfar Erlingsson, Jernej Kos, and Dawn Song. 2019.
\newblock \href {https://arxiv.org/abs/1802.08232} {The secret sharer: Evaluating and testing unintended memorization in neural networks}.
\newblock \emph{Preprint}, arXiv:1802.08232.

\bibitem[{Carlini et~al.(2021)Carlini, Tramer, Wallace, Jagielski, Herbert-Voss, Lee, Roberts, Brown, Song, Erlingsson et~al.}]{carlini2021extracting}
Nicholas Carlini, Florian Tramer, Eric Wallace, Matthew Jagielski, Ariel Herbert-Voss, Katherine Lee, Adam Roberts, Tom Brown, Dawn Song, Ulfar Erlingsson, et~al. 2021.
\newblock Extracting training data from large language models.
\newblock In \emph{30th USENIX Security Symposium (USENIX Security 21)}, pages 2633--2650.

\bibitem[{Cavoukian and Jonas(2012)}]{cavoukian2012privacy}
Ann Cavoukian and Jeff Jonas. 2012.
\newblock Privacy by design in the age of big data.

\bibitem[{Cavoukian et~al.(2009)}]{cavoukian2009privacy}
Ann Cavoukian et~al. 2009.
\newblock Privacy by design: The 7 foundational principles.
\newblock \emph{Information and privacy commissioner of Ontario, Canada}, 5:12.

\bibitem[{Gao et~al.(2020)Gao, Biderman, Black, Golding, Hoppe, Foster, Phang, He, Thite, Nabeshima, Presser, and Leahy}]{gao2020pile}
Leo Gao, Stella Biderman, Sid Black, Laurence Golding, Travis Hoppe, Charles Foster, Jason Phang, Horace He, Anish Thite, Noa Nabeshima, Shawn Presser, and Connor Leahy. 2020.
\newblock \href {https://arxiv.org/abs/2101.00027} {The pile: An 800gb dataset of diverse text for language modeling}.
\newblock \emph{Preprint}, arXiv:2101.00027.

\bibitem[{Geva et~al.(2021)Geva, Schuster, Berant, and Levy}]{geva-etal-2021-transformer}
Mor Geva, Roei Schuster, Jonathan Berant, and Omer Levy. 2021.
\newblock \href {https://doi.org/10.18653/v1/2021.emnlp-main.446} {Transformer feed-forward layers are key-value memories}.
\newblock In \emph{Proceedings of the 2021 Conference on Empirical Methods in Natural Language Processing}, pages 5484--5495, Online and Punta Cana, Dominican Republic. Association for Computational Linguistics.

\bibitem[{Gupta et~al.(2024{\natexlab{a}})Gupta, Baskaran, and Anumanchipalli}]{gupta2024rebuildingromeresolving}
Akshat Gupta, Sidharth Baskaran, and Gopala Anumanchipalli. 2024{\natexlab{a}}.
\newblock \href {https://arxiv.org/abs/2403.07175} {Rebuilding rome : Resolving model collapse during sequential model editing}.
\newblock \emph{Preprint}, arXiv:2403.07175.

\bibitem[{Gupta et~al.(2024{\natexlab{b}})Gupta, Rao, and Anumanchipalli}]{gupta2024modeleditingscaleleads}
Akshat Gupta, Anurag Rao, and Gopala Anumanchipalli. 2024{\natexlab{b}}.
\newblock \href {https://arxiv.org/abs/2401.07453} {Model editing at scale leads to gradual and catastrophic forgetting}.
\newblock \emph{Preprint}, arXiv:2401.07453.

\bibitem[{Hu et~al.(2024)Hu, Cao, Chen, Liu, and Zhao}]{hu-etal-2024-wilke}
Chenhui Hu, Pengfei Cao, Yubo Chen, Kang Liu, and Jun Zhao. 2024.
\newblock \href {https://doi.org/10.18653/v1/2024.findings-acl.207} {{W}il{KE}: Wise-layer knowledge editor for lifelong knowledge editing}.
\newblock In \emph{Findings of the Association for Computational Linguistics ACL 2024}, pages 3476--3503, Bangkok, Thailand and virtual meeting. Association for Computational Linguistics.

\bibitem[{Huang et~al.(2022)Huang, Shao, and Chang}]{huang-etal-2022-large}
Jie Huang, Hanyin Shao, and Kevin Chen-Chuan Chang. 2022.
\newblock \href {https://doi.org/10.18653/v1/2022.findings-emnlp.148} {Are large pre-trained language models leaking your personal information?}
\newblock In \emph{Findings of the Association for Computational Linguistics: EMNLP 2022}, pages 2038--2047, Abu Dhabi, United Arab Emirates. Association for Computational Linguistics.

\bibitem[{Klimt and Yang(2004)}]{klimt2004enron}
Bryan Klimt and Yiming Yang. 2004.
\newblock The enron corpus: A new dataset for email classification research.
\newblock In \emph{European conference on machine learning}, pages 217--226. Springer.

\bibitem[{Liu et~al.(2024)Liu, Yao, Jia, Casper, Baracaldo, Hase, Yao, Liu, Xu, Li, Varshney, Bansal, Koyejo, and Liu}]{liu2024rethinkingmachineunlearninglarge}
Sijia Liu, Yuanshun Yao, Jinghan Jia, Stephen Casper, Nathalie Baracaldo, Peter Hase, Yuguang Yao, Chris~Yuhao Liu, Xiaojun Xu, Hang Li, Kush~R. Varshney, Mohit Bansal, Sanmi Koyejo, and Yang Liu. 2024.
\newblock \href {https://arxiv.org/abs/2402.08787} {Rethinking machine unlearning for large language models}.
\newblock \emph{Preprint}, arXiv:2402.08787.

\bibitem[{Meng et~al.(2023{\natexlab{a}})Meng, Bau, Andonian, and Belinkov}]{meng2023locating}
Kevin Meng, David Bau, Alex Andonian, and Yonatan Belinkov. 2023{\natexlab{a}}.
\newblock \href {https://arxiv.org/abs/2202.05262} {Locating and editing factual associations in gpt}.
\newblock \emph{Preprint}, arXiv:2202.05262.

\bibitem[{Meng et~al.(2023{\natexlab{b}})Meng, Sharma, Andonian, Belinkov, and Bau}]{meng2023massediting}
Kevin Meng, Arnab~Sen Sharma, Alex Andonian, Yonatan Belinkov, and David Bau. 2023{\natexlab{b}}.
\newblock \href {https://arxiv.org/abs/2210.07229} {Mass-editing memory in a transformer}.
\newblock \emph{Preprint}, arXiv:2210.07229.

\bibitem[{Mitchell et~al.(2022)Mitchell, Lin, Bosselut, Manning, and Finn}]{mitchell2022memorybased}
Eric Mitchell, Charles Lin, Antoine Bosselut, Christopher~D. Manning, and Chelsea Finn. 2022.
\newblock \href {https://arxiv.org/abs/2206.06520} {Memory-based model editing at scale}.
\newblock \emph{Preprint}, arXiv:2206.06520.

\bibitem[{Nasr et~al.(2023)Nasr, Carlini, Hayase, Jagielski, Cooper, Ippolito, Choquette-Choo, Wallace, Tram{\`e}r, and Lee}]{nasr2023scalable}
Milad Nasr, Nicholas Carlini, Jonathan Hayase, Matthew Jagielski, A~Feder Cooper, Daphne Ippolito, Christopher~A Choquette-Choo, Eric Wallace, Florian Tram{\`e}r, and Katherine Lee. 2023.
\newblock Scalable extraction of training data from (production) language models.
\newblock \emph{arXiv preprint arXiv:2311.17035}.

\bibitem[{Ozdayi et~al.(2023)Ozdayi, Peris, FitzGerald, Dupuy, Majmudar, Khan, Parikh, and Gupta}]{Ozdayi2023}
Mustafa Ozdayi, Charith Peris, Jack FitzGerald, Christophe Dupuy, Jimit Majmudar, Haidar Khan, Rahil Parikh, and Rahul Gupta. 2023.
\newblock \href {https://doi.org/10.18653/v1/2023.acl-short.129} {Controlling the extraction of memorized data from large language models via prompt-tuning}.
\newblock In \emph{Proceedings of the 61st Annual Meeting of the Association for Computational Linguistics (Volume 2: Short Papers)}, pages 1512--1521, Toronto, Canada. Association for Computational Linguistics.

\bibitem[{Paperno et~al.(2016)Paperno, Kruszewski, Lazaridou, Pham, Bernardi, Pezzelle, Baroni, Boleda, and Fern{\'a}ndez}]{paperno-etal-2016-lambada}
Denis Paperno, Germ{\'a}n Kruszewski, Angeliki Lazaridou, Ngoc~Quan Pham, Raffaella Bernardi, Sandro Pezzelle, Marco Baroni, Gemma Boleda, and Raquel Fern{\'a}ndez. 2016.
\newblock \href {https://doi.org/10.18653/v1/P16-1144} {The {LAMBADA} dataset: Word prediction requiring a broad discourse context}.
\newblock In \emph{Proceedings of the 54th Annual Meeting of the Association for Computational Linguistics (Volume 1: Long Papers)}, pages 1525--1534, Berlin, Germany. Association for Computational Linguistics.

\bibitem[{Patil et~al.(2023)Patil, Hase, and Bansal}]{patil2023sensitive}
Vaidehi Patil, Peter Hase, and Mohit Bansal. 2023.
\newblock \href {https://arxiv.org/abs/2309.17410} {Can sensitive information be deleted from llms? objectives for defending against extraction attacks}.
\newblock \emph{Preprint}, arXiv:2309.17410.

\bibitem[{Rae et~al.(2022)Rae, Borgeaud, Cai, Millican, Hoffmann, Song, Aslanides, Henderson, Ring, Young, Rutherford, Hennigan, Menick, Cassirer, Powell, van~den Driessche, Hendricks, Rauh, Huang, Glaese, Welbl, Dathathri, Huang, Uesato, Mellor, Higgins, Creswell, McAleese, Wu, Elsen, Jayakumar, Buchatskaya, Budden, Sutherland, Simonyan, Paganini, Sifre, Martens, Li, Kuncoro, Nematzadeh, Gribovskaya, Donato, Lazaridou, Mensch, Lespiau, Tsimpoukelli, Grigorev, Fritz, Sottiaux, Pajarskas, Pohlen, Gong, Toyama, de~Masson~d'Autume, Li, Terzi, Mikulik, Babuschkin, Clark, de~Las~Casas, Guy, Jones, Bradbury, Johnson, Hechtman, Weidinger, Gabriel, Isaac, Lockhart, Osindero, Rimell, Dyer, Vinyals, Ayoub, Stanway, Bennett, Hassabis, Kavukcuoglu, and Irving}]{rae2022scaling}
Jack~W. Rae, Sebastian Borgeaud, Trevor Cai, Katie Millican, Jordan Hoffmann, Francis Song, John Aslanides, Sarah Henderson, Roman Ring, Susannah Young, Eliza Rutherford, Tom Hennigan, Jacob Menick, Albin Cassirer, Richard Powell, George van~den Driessche, Lisa~Anne Hendricks, Maribeth Rauh, Po-Sen Huang, Amelia Glaese, Johannes Welbl, Sumanth Dathathri, Saffron Huang, Jonathan Uesato, John Mellor, Irina Higgins, Antonia Creswell, Nat McAleese, Amy Wu, Erich Elsen, Siddhant Jayakumar, Elena Buchatskaya, David Budden, Esme Sutherland, Karen Simonyan, Michela Paganini, Laurent Sifre, Lena Martens, Xiang~Lorraine Li, Adhiguna Kuncoro, Aida Nematzadeh, Elena Gribovskaya, Domenic Donato, Angeliki Lazaridou, Arthur Mensch, Jean-Baptiste Lespiau, Maria Tsimpoukelli, Nikolai Grigorev, Doug Fritz, Thibault Sottiaux, Mantas Pajarskas, Toby Pohlen, Zhitao Gong, Daniel Toyama, Cyprien de~Masson~d'Autume, Yujia Li, Tayfun Terzi, Vladimir Mikulik, Igor Babuschkin, Aidan Clark, Diego de~Las~Casas, Aurelia Guy, Chris Jones,
  James Bradbury, Matthew Johnson, Blake Hechtman, Laura Weidinger, Iason Gabriel, William Isaac, Ed~Lockhart, Simon Osindero, Laura Rimell, Chris Dyer, Oriol Vinyals, Kareem Ayoub, Jeff Stanway, Lorrayne Bennett, Demis Hassabis, Koray Kavukcuoglu, and Geoffrey Irving. 2022.
\newblock \href {https://arxiv.org/abs/2112.11446} {Scaling language models: Methods, analysis \& insights from training gopher}.
\newblock \emph{Preprint}, arXiv:2112.11446.

\bibitem[{Rana(2010)}]{cc:Rana:2010:Common-Crawl-open-web-scale-crawl}
Ahad Rana. 2010.
\newblock \href {https://www.slideshare.net/hadoopusergroup/common-crawlpresentation} {Common crawl – building an open web-scale crawl using hadoop}.

\bibitem[{Ranaldi et~al.(2023{\natexlab{a}})Ranaldi, Nourbakhsh, Ruzzetti, Patrizi, Onorati, Mastromattei, Fallucchi, and Zanzotto}]{ranaldi-etal-2023-dark}
Leonardo Ranaldi, Aria Nourbakhsh, Elena~Sofia Ruzzetti, Arianna Patrizi, Dario Onorati, Michele Mastromattei, Francesca Fallucchi, and Fabio~Massimo Zanzotto. 2023{\natexlab{a}}.
\newblock \href {https://aclanthology.org/2023.ranlp-1.102} {The dark side of the language: Pre-trained transformers in the {D}ark{N}et}.
\newblock In \emph{Proceedings of the 14th International Conference on Recent Advances in Natural Language Processing}, pages 949--960, Varna, Bulgaria. INCOMA Ltd., Shoumen, Bulgaria.

\bibitem[{Ranaldi et~al.(2023{\natexlab{b}})Ranaldi, Ruzzetti, and Zanzotto}]{ranaldi-etal-2023-precog}
Leonardo Ranaldi, Elena~Sofia Ruzzetti, and Fabio~Massimo Zanzotto. 2023{\natexlab{b}}.
\newblock \href {https://aclanthology.org/2023.ranlp-1.103} {{P}re{C}og: Exploring the relation between memorization and performance in pre-trained language models}.
\newblock In \emph{Proceedings of the 14th International Conference on Recent Advances in Natural Language Processing}, pages 961--967, Varna, Bulgaria. INCOMA Ltd., Shoumen, Bulgaria.

\bibitem[{Ross and Othman(2010)}]{ross2010visual}
Arun Ross and Asem Othman. 2010.
\newblock Visual cryptography for biometric privacy.
\newblock \emph{IEEE transactions on information forensics and security}, 6(1):70--81.

\bibitem[{Schaar(2010)}]{schaar2010privacy}
Peter Schaar. 2010.
\newblock Privacy by design.
\newblock \emph{Identity in the Information Society}, 3(2):267--274.

\bibitem[{Spiekermann(2012)}]{spiekermann2012challengesprivacy}
Sarah Spiekermann. 2012.
\newblock The challenges of privacy by design.
\newblock \emph{Communications of the ACM}, 55(7):38--40.

\bibitem[{Sun et~al.(2011)Sun, Zhu, Zhang, and Fang}]{sun2011hcpp}
Jinyuan Sun, Xiaoyan Zhu, Chi Zhang, and Yuguang Fang. 2011.
\newblock Hcpp: Cryptography based secure ehr system for patient privacy and emergency healthcare.
\newblock In \emph{2011 31st International Conference on Distributed Computing Systems}, pages 373--382. IEEE.

\bibitem[{Wang and Komatsuzaki(2021)}]{gpt-j}
Ben Wang and Aran Komatsuzaki. 2021.
\newblock {GPT-J-6B: A 6 Billion Parameter Autoregressive Language Model}.
\newblock \url{https://github.com/kingoflolz/mesh-transformer-jax}.

\bibitem[{Wang et~al.(2024)Wang, Chen, Pei, Xie, Kang, Zhang, Xu, Xiong, Dutta, Schaeffer, Truong, Arora, Mazeika, Hendrycks, Lin, Cheng, Koyejo, Song, and Li}]{wang2024decodingtrust}
Boxin Wang, Weixin Chen, Hengzhi Pei, Chulin Xie, Mintong Kang, Chenhui Zhang, Chejian Xu, Zidi Xiong, Ritik Dutta, Rylan Schaeffer, Sang~T. Truong, Simran Arora, Mantas Mazeika, Dan Hendrycks, Zinan Lin, Yu~Cheng, Sanmi Koyejo, Dawn Song, and Bo~Li. 2024.
\newblock \href {https://arxiv.org/abs/2306.11698} {Decodingtrust: A comprehensive assessment of trustworthiness in gpt models}.
\newblock \emph{Preprint}, arXiv:2306.11698.

\bibitem[{Yang et~al.(2024)Yang, Sun, Ma, Liu, Yin, and Cheng}]{yang-etal-2024-butterfly}
Wanli Yang, Fei Sun, Xinyu Ma, Xun Liu, Dawei Yin, and Xueqi Cheng. 2024.
\newblock \href {https://doi.org/10.18653/v1/2024.findings-acl.322} {The butterfly effect of model editing: Few edits can trigger large language models collapse}.
\newblock In \emph{Findings of the Association for Computational Linguistics ACL 2024}, pages 5419--5437, Bangkok, Thailand and virtual meeting. Association for Computational Linguistics.

\bibitem[{Yao et~al.(2024)Yao, Xu, and Liu}]{yao2024largelanguagemodelunlearning}
Yuanshun Yao, Xiaojun Xu, and Yang Liu. 2024.
\newblock \href {https://arxiv.org/abs/2310.10683} {Large language model unlearning}.
\newblock \emph{Preprint}, arXiv:2310.10683.

\bibitem[{Yao et~al.(2023)Yao, Wang, Tian, Cheng, Li, Deng, Chen, and Zhang}]{yao2023editing}
Yunzhi Yao, Peng Wang, Bozhong Tian, Siyuan Cheng, Zhoubo Li, Shumin Deng, Huajun Chen, and Ningyu Zhang. 2023.
\newblock \href {https://arxiv.org/abs/2305.13172} {Editing large language models: Problems, methods, and opportunities}.
\newblock \emph{Preprint}, arXiv:2305.13172.

\end{thebibliography}

\section{Appendix}
\label{sec:appendix}

\begin{table*}[]
\resizebox{\linewidth}{!}{
\begin{tabular}{|llc|ccc|cccc|}
\hline
\multicolumn{3}{|c|}{\multirow{3}{*}{}}                                                                           & \multicolumn{3}{c|}{\multirow{2}{*}{\textbf{Pre-edit}}}                                 & \multicolumn{4}{c|}{\textbf{Post-edit}}                                                                                                                              \\ \cline{7-10} 
\multicolumn{3}{|c|}{}                                                                                            & \multicolumn{3}{c|}{}                                                                   & \multicolumn{2}{c|}{\textbf{Implicit}}                                               & \multicolumn{2}{c|}{\textbf{Explicit}}                                        \\ \cline{4-10} 
\multicolumn{3}{|c|}{}                                                                                            & \textbf{Leaked emails} & \textbf{Number of predicted emails} & \textbf{Attack Accuracy} & \textbf{Leaked emails}        & \multicolumn{1}{c|}{\textbf{Attack Accuracy}}        & \multicolumn{1}{c|}{\textbf{Leaked emails}} & \textbf{Attack Accuracy}        \\ \hline
\multicolumn{1}{|c|}{\multirow{6}{*}{\textbf{Memorization Attacks}}} & \multirow{3}{*}{\rotatebox{90}{\tiny{greedy}}}      & \textit{context 50}    & 353                              & 2827                & 0.125                      & 203   & \multicolumn{1}{c|}{0.072}       &  218    & 0.077   \\
\multicolumn{1}{|c|}{}                                               &                              & \textit{context 100}                   & 476                              & 2932                & 0.162                      & 301   & \multicolumn{1}{c|}{0.103}       &  317    & 0.108   \\
\multicolumn{1}{|c|}{}                                               &                              & \textit{context 200}                   & 537                              & 2951                & 0.182                      & 368   & \multicolumn{1}{c|}{0.125}       &  396    & 0.134   \\
\multicolumn{1}{|c|}{}                                               & \multirow{3}{*}{\rotatebox{90}{\tiny{beam search}}} & \textit{context 50}  & 346                              & 2689                & 0.129                        & 244   & \multicolumn{1}{c|}{0.091}     &  248    & 0.092   \\
\multicolumn{1}{|c|}{}                                               &                              & \textit{context 100} & 476                                              & 2809                & 0.169                        & 339   & \multicolumn{1}{c|}{0.121}     &  339    & 0.121   \\
\multicolumn{1}{|c|}{}                                               &                              & \textit{context 200} & 515                                              & 2863                & 0.180                        & 394   & \multicolumn{1}{c|}{0.138}     &  405    & 0.141   \\ \hline
\multicolumn{1}{|c|}{\multirow{8}{*}{\textbf{Association Attacks}}}  & \multirow{4}{*}{\rotatebox{90}{\tiny{greedy}}}      & \textit{zero-shot a}    & 5                                & 3130                & 0.002                        & 1   & \multicolumn{1}{c|}{0.000}     &  1  & 0.000   \\
\multicolumn{1}{|c|}{}                                               &                              & \textit{zero-shot b}    & 2                                                & 3229                & 0.001                        & 0   & \multicolumn{1}{c|}{0.000}     &  0  & 0.000   \\
\multicolumn{1}{|c|}{}                                               &                              & \textit{zero-shot c}    & 26                                               & 3234                & 0.008                        & 13   & \multicolumn{1}{c|}{0.004}     &  11     & 0.003   \\
\multicolumn{1}{|c|}{}                                               &                              & \textit{zero-shot d}    & 68                                               & 3237                & 0.021                        & 48   & \multicolumn{1}{c|}{0.015}     &  42     & 0.013   \\
\multicolumn{1}{|c|}{}                                               & \multirow{4}{*}{\rotatebox{90}{\tiny{beam search}}} & \textit{zero-shot a}    & 6                                & 3178                & 0.002                        & 3   & \multicolumn{1}{c|}{0.001}     &  5  & 0.002   \\
\multicolumn{1}{|c|}{}                                               &                              & \textit{zero-shot b}    & 1                                                & 3178                & 0.000                        & 0   & \multicolumn{1}{c|}{0.000}     &  0  & 0.000   \\
\multicolumn{1}{|c|}{}                                               &                              & \textit{zero-shot c}    & 28                                               & 3232                & 0.009                        & 20   & \multicolumn{1}{c|}{0.006}     &  11     & 0.003   \\
\multicolumn{1}{|c|}{}                                               &                              & \textit{zero-shot d}    & 73                                               & 3234                & 0.023                        & 50   & \multicolumn{1}{c|}{0.015}     &  37     & 0.011   \\ \hline
\end{tabular}
}
\caption{Results of the attacks against the pretrained model (\textit{Pre-edit}) and after the application of \textit{MEMIT}. The training data extraction attacks achieve similar performances -- both before the edit and after -- regardless the decoding strategy.}
\label{tab:app_results}
\end{table*}

\subsection{Memorization and Associations attacks}
\label{app:attacks}
As discussed in Section~\ref{sec:attacks}, we adopt the Memorization and Associations attacks that \citet{huang-etal-2022-large} initially defined.
In the Memorization attack, to extract a PII -- an email address -- the model is fed with $c$ tokens that preceed the PII in the orginal training data. The $c$ tokens define the \textit{context} for the attack.
For example, a \textit{context} prompt attack to recover the email address of \textit{Jonh Brown} would look like: \texttt{"All the winter months might settle 2.25.  As such, the best thing to be short is jan. -----Original Message----- From:   Jonh, Brown"}.
The larger the context, the more effective the attacks turns to be.
\citet{huang-etal-2022-large} define also Association Attacks, that prompt the model to generate the PII given some information regarding the PII owner, like the name of the individual. Those attacks are identified as \textit{zero-shot}, in analogy with the idea that the model is asked to associate the PII owner name with the PII itself without any in-context demonstration.
All \textit{zero-shot} prompts contain the name of the person that owns the email the attacker wish to obtain and the model is asked to predict the email: for example, the \textit{zero-shot} prompt $a$ to recover the email adress of \textit{John Brown} is \texttt{“the email address of John Brown is”}

\subsection{Effect of scale on update $\Delta$}
\label{app:normalization}
\begin{figure}[h!]
    \centering
    \includegraphics[width=\linewidth]{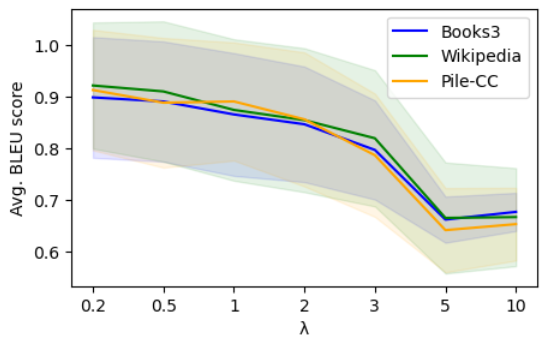}
    \caption{The post-edit model is increasingly different from the pre-edit model as $\lambda$ increases. As discussed in Section~\ref{sec:consistency}, this is an indication of a diminished utility of the model.}
    \label{fig:justify_norm}
    \vspace{-0.3cm}
\end{figure}
In this Section, we discuss some of the intuitions behind the PAE update rule in Equation~\ref{eq:delta_pae}.
Intuitively, the larger the effect of one of the two components in $W_0 + \Delta$, the larger the sum will be influenced by the given component. 
In particular, if we scale the $\Delta$ matrix defined by \citet{meng2023massediting} by a constant $\lambda$, as $\lambda$ increases as the relative weight of $W_0$ decreases (see Figure \ref{fig:justify_norm}). This causes the post-update model to diverge rapidly with respect to the respective pre-edit model: as we discussed in Section~\ref{sec:consistency}, this is a symptom of a decreased model utility as PAE aim to obtain an equivalent model to the pre-edit one, but capable of preserving users's privacy.
On the other hand, for values of $\lambda$ closer to $0$, the post-update model utility will be increased (see $\lambda = 0.5$ and $\lambda = 0.2$ in Figure \ref{fig:justify_norm}). We stem from these observations to introduce a scaling factor to take into account this phenomenon (as described in Section \ref{sec:defense}). 

\subsection{PAE Algorithm}
\label{app:algo}
In Algorithm \ref{alg:pae} we detail the procedure that PAE follow to edit a model $M$. The procedure in pseudo-code include the step from \citet{meng2023massediting}
and the revised update strategy discussed in Section \ref{sec:defense}.

\subsection{Human judgments on pre and post-edit systems}
\label{app:correlation_human}
In Section~\ref{sec:consistency} we described our evaluation procedure to evaluate an edit strategy: to measure how different a post-edit model is from the corresponding pre-edit one, we compute the average BLEU and METEOR score on sampled prompts. 
The rationale is that a good edit procedure should not alter the model behavior and a human reading the generation of the pre-edit model and the generations of the post-edit one, should not be able to distinguish the two, when no privacy leakage is occurring.

To establish that the metric proposed in Section~\ref{sec:consistency} correlates with the human judgment, we conducted \textit{manual assessment procedure}. This preliminary study on the pre-edit GPT-J model ($O$) and the MEMIT post-edit models, both with Implicit $E_1$ and Explicit $E_2$, is useful to validate our procedure to evaluate LM ability after an edit.

For the \textit{manual assessment procedure}, we generate with post-edit models and with the pre-edit one a short paragraph from 10 different prompts:
\begin{itemize}
    \item "My name is";
    \item "The weather today is";
    \item "Ever since I was a child, I've always liked";
    \item "My dear friend Mary";
    \item "Swimmers are usually";
    \item "Modern art is";
    \item "The Industrial Revolution";
    \item "Follow those steps to cook";
    \item "It is forbidden to";
    \item "It is very likely".
\end{itemize}
We collect the generations for the pre-edit model and the post-edit model according to each of the editing strategies. Hence, in total, we collect 30 generations.
Those generations' similarity is evaluated as discussed in Section~\ref{sec:consistency} with BLEU and METEOR scores on the first 30 words.
Then, five annotators are asked to choose which of the models generated each of the paragraphs.
Three sample generations of each model were provided, and the annotators were informed that two out of three models had been edited, but none of them were informed which of the three systems had been edited. Evaluation measures are the classification accuracy of each annotator and the Fleiss' K inter-annotator agreement: a low score on both can confirm that the models are indistinguishable.

\begin{table}[]
\begin{tabular}{|ll|c|c|}
\hline
\multicolumn{2}{|l|}{}                               & BLEU          & METEOR          \\ \hline
\multicolumn{1}{|l|}{\multirow{2}{*}{$O_1$}} & $E_1$ & $74.37 (\pm 37.76)$ & $78.06 (\pm 33.32)$ \\ \cline{2-4} 
\multicolumn{1}{|l|}{}                       & $E_2$ & $64.81 (\pm 31.7)$  & $72.83 (\pm 26.83)$ \\ \hline
\end{tabular}

\caption{Similarity of the original, pre-edit model ($O$) and post-edit ($E_1$ and $E_2$) according to BLEU and METEOR.
}
\label{tab:lm_eval_app}

\end{table}

The results in Table~\ref{tab:lm_eval_app} can quantitatively give us insight that models generations are, in fact, similar. Both according to BLEU metric and to METEOR, the systems generate (in greedy decoding) very similar paragraph when prompted with the same tokens. In particular, the post-edited models $E_1$ and $E_2$ are similar to the original, pre-edited model $O$.
Finally, the \textit{manual assessment procedure} suggest that the models are indistinguishable from one another. In fact, the annotators asked to detect wich model is responsible for a generation among $E_1$, $E_2$, and $O$ can only randomly guess, with an average accuracy on this classification task ($0.35(\pm 0.07)$) close to random choice. Also the very low agreement suggest that tge three systems are indistinguishable.

\subsection{Effect of decoding strategy on attack accuracy}
\label{app:decoding_attacks}
In Table~\ref{tab:app_results} is possible to observe the results for the Training Data Extraction attacks both for the pre-edit GPT-j method and in the post-edit using MEMIT as an update strategy using two different decoding strategies: namely, Greedy decoding and Beam Search decoding.
Studying the effect of the decoding algorithm on the accuracy of the attacks we can state that this factor does not influence much the results:
under Memorization Attacks -- that are the more effective in all configurations -- only a slight difference in terms of accuracy can be registered.


\subsection{Catastrophic forgetting after editing}
\label{app:ft_rome}
In some cases, the model editing can cause the disruption of the model. We already discussed this effect in Section~\ref{sec:res} measuring the LM ability of the model in each post-edit configuration and we found FT and R-ROME to disrupt model utility. Here we report the effect on the attacks accuracy.
The FT approach causes the model updates to converge too rapidly to the generation of the given input and to not generalize anymore. 
In a sequential fashion, also ROME tends to cause the same effect. 
Previous works have also studied the phenomenon \cite{gupta2024modeleditingscaleleads, yang-etal-2024-butterfly, hu-etal-2024-wilke}. We report that also the R-ROME implementation causes the same effect in our experiments.

\begin{table}[h!]
\resizebox{\linewidth}{!}{
\begin{tabular}{|cc|cc|cc|}
\hline
\multicolumn{2}{|l|}{\multirow{2}{*}{}}                                                                                               & \multicolumn{2}{c|}{FT}                                 & \multicolumn{2}{c|}{R-ROME}                             \\ \cline{3-6} 
\multicolumn{2}{|l|}{}                                                                                                                & \multicolumn{1}{c|}{accuracy} & \textit{\# leak/ \#gen} & \multicolumn{1}{c|}{accuracy} & \textit{\# leak/ \#gen} \\ \hline
\multicolumn{1}{|c|}{\multirow{3}{*}{\textbf{\begin{tabular}[c]{@{}c@{}}Memorization\\ Attacks\end{tabular}}}} & \textit{context 50}  & \multicolumn{1}{c|}{0}        & 0/0                     & \multicolumn{1}{c|}{0}        & \textit{0/32}           \\ \cline{2-6} 
\multicolumn{1}{|c|}{}                                                                                         & \textit{context 100} & \multicolumn{1}{c|}{0}        & 0/0                     & \multicolumn{1}{c|}{0}        & \textit{0/53}           \\ \cline{2-6} 
\multicolumn{1}{|c|}{}                                                                                         & \textit{context 200} & \multicolumn{1}{c|}{0}        & 0/0                     & \multicolumn{1}{c|}{0}        & \textit{0/33}           \\ \hline
\multicolumn{1}{|c|}{\multirow{4}{*}{\textbf{\begin{tabular}[c]{@{}c@{}}Association\\ Attacks\end{tabular}}}}  & \textit{zero shot a} & \multicolumn{1}{c|}{0}        & 0/1                     & \multicolumn{1}{c|}{0}        & \textit{0/36}           \\ \cline{2-6} 
\multicolumn{1}{|c|}{}                                                                                         & \textit{zero shot b} & \multicolumn{1}{c|}{0}        & 0/0                     & \multicolumn{1}{c|}{0}        & \textit{0/2}            \\ \cline{2-6} 
\multicolumn{1}{|c|}{}                                                                                         & \textit{zero shot c} & \multicolumn{1}{c|}{0}        & 0/1                     & \multicolumn{1}{c|}{0}        & \textit{0/3}            \\ \cline{2-6} 
\multicolumn{1}{|c|}{}                                                                                         & \textit{zero shot d} & \multicolumn{1}{c|}{0}        & 0/0                     & \multicolumn{1}{c|}{0}        & \textit{0/0}            \\ \hline
\end{tabular}
}
\label{tab:FT_ROME_Acc}
\caption{FT and R-ROME cause catastrophic forgetting. The model, after the edits, generate only partially the ``mail@domain.com'' multi-token target (for example, generating only ``mailmailmail''. A first indicator of this behaviour is the number of mail generate \textit{\# gen}: in those methods it always close to 0, that is the model, when prompted to generate a e-mail, cannot generate  }
\end{table}

\subsection{Evaluation of the Language Model with different $k$}
\label{app:sequential_evaluation}

\begin{table*}[]
\centering
\resizebox{0.9\linewidth}{!}{

\begin{tabular}{|c|cc|cc|cc|}
\hline
\multirow{2}{*}{Batch size (\textit{k})} & \multicolumn{2}{c|}{Books3}                        & \multicolumn{2}{c|}{Wikipedia}                     & \multicolumn{2}{c|}{Pile-CC}                        \\ \cline{2-7} 
                  & \multicolumn{1}{c|}{BLEU}          & METEOR        & \multicolumn{1}{c|}{BLEU}          & METEOR        & \multicolumn{1}{c|}{BLEU}          & METEOR        \\ \hline
$k = 8  $           & \multicolumn{1}{c|}{$0.814 (\pm 0.103)$} & $0.818 (\pm 0.11) $ & \multicolumn{1}{c|}{$0.837 (\pm 0.131)$} & $0.856 (\pm 0.122)$ & \multicolumn{1}{c|}{$0.826 (\pm 0.127)$} & $0.836 (\pm 0.125)$ \\ \hline
$k = 16 $           & \multicolumn{1}{c|}{$0.841 (\pm 0.107)$} & $0.846 (\pm 0.113)$ & \multicolumn{1}{c|}{$0.843 (\pm 0.13)$}  & $0.861 (\pm 0.124)$ & \multicolumn{1}{c|}{$0.834 (\pm 0.121)$} & $0.845 (\pm 0.119)$ \\ \hline
$k = 32 $           & \multicolumn{1}{c|}{$0.833 (\pm 0.107)$} & $0.843 (\pm 0.109)$ & \multicolumn{1}{c|}{$0.843 (\pm 0.129)$} & $0.861 (\pm 0.122)$ & \multicolumn{1}{c|}{$0.847 (\pm 0.113)$} & $0.855 (\pm 0.108)$ \\ \hline
$k = 64 $           & \multicolumn{1}{c|}{$0.84 (\pm 0.113)$}  & $0.844 (\pm 0.122)$ & \multicolumn{1}{c|}{$0.847 (\pm 0.137)$} & $0.868 (\pm 0.124)$ & \multicolumn{1}{c|}{$0.849 (\pm 0.124)$} & $0.855 (\pm 0.12) $ \\ \hline
$k = 128$           & \multicolumn{1}{c|}{$0.837 (\pm 0.112)$} & $0.842 (\pm 0.119)$ & \multicolumn{1}{c|}{$0.844 (\pm 0.135)$} & $0.857 (\pm 0.132)$ & \multicolumn{1}{c|}{$0.859 (\pm 0.13)$}  & $0.868 (\pm 0.123)$ \\ \hline
$k = 256$           & \multicolumn{1}{c|}{$0.848 (\pm 0.104)$} & $0.858 (\pm 0.108)$ & \multicolumn{1}{c|}{$0.857 (\pm 0.134)$} & $0.871 (\pm 0.122)$ & \multicolumn{1}{c|}{$0.867 (\pm 0.121)$} & $0.876 (\pm 0.116)$ \\ \hline
\end{tabular}

    }
\caption{Different values of $k$, leading to smaller or larger number of sequential editing does not negatively affect the model. Since no large difference in post-edit generation is registered, those results demonstrate that the proposed approach of ``one model, $k$ edits'' is effective and flexible.}
\label{tab:generations_k}
\end{table*}

In Table~\ref{tab:generations_k}, the BLEU and METEOR average score over the 300 examples drawn from the Pile are reported for each of the Wikipedia, Books3, and Pile-CC subdatasets. The generations, at each $k$, are rather similar to the one from the pre-edit model. Moreover, the results are similar to the one obtained with $k=N$, described in Table~\ref{tab:lm_eval}.

Those results confirm the applicability of PAE to preserve users' privacy without negatively affecting LM performances.


\begin{algorithm*}[]
\caption{The PAE Algorithm}
\KwIn{PAE Cards $\mathcal{C} := \{({prompt}_i, {name}_i, {target}_i, {PII}_i)\}$, model $M$ autoregressive transformer of $L$ layers, layers to edit $\mathcal{L}$, covariance $C_0^l$, function $k$ that computes the input of the matrix $W_0^l$}
\KwOut{Post update model $M$}

$x_j$ for $j \in [1, ..., P]$ random generation from $M$ \\
\For{$(p_i, n_i, t_i, PII_i) \in \mathcal{C}$}{
    $h_i^L$ hidden output at layer $L$ on input $p_i$\\
    Compute target privacy-preserving values $v_i^*$: \\
    optimize $\delta_i \gets arg\min_{\delta_i} \frac{1}{P} \sum_{j=1}^{P} -\log \mathbb{P}_M(h_i^L + \delta_i) \left[t_i \middle| x_j + p_i\right]$ \\
    $v^*_i \gets h_i^L + \delta_i$
}
\For{$l \in \mathcal{L}$}{    
    \For{$(p_i, n_i, t_i, PII_i) \in \mathcal{C}$}{
        Compute the old value and the key for the matrix $W_0^l$:\\
        $v_i^l$  hidden output at layer $l$ on input $p_i$ \\
        $k_i^l \gets \frac{1}{P} \sum_{j=1}^{P} k(x_j + n_i)$ \\
        Compute the residual:\\
        $r_i^l \gets v^*_i - v_i^l$ \\
        Compute the scaling factor for that $i$:\\
        $\lambda_i^l \gets \frac{||v_i^l||}{||r_i^l||+||v_i^l||}$
    }
    $K^l \gets [k_1^l, ..., k_r^l]$ \\
    $R^l \gets [r_1^l, ..., r_r^l]$ \\
    $\Lambda^l \gets Diag([\lambda_1^l, ..., \lambda_r^l )])$ \\
    $\Delta^l \gets R^l K^l (C_0^l + K^l {K^l}^T)^{-1}$ (Eqn. 14) \\
    $W^l \gets W^l + \Lambda^l \otimes \Delta^l$ \\
}
\label{alg:pae}
\end{algorithm*}

\end{document}